\PassOptionsToPackage{unicode}{hyperref}
\PassOptionsToPackage{hyphens}{url}
\PassOptionsToPackage{dvipsnames,svgnames,x11names}{xcolor}
\documentclass[
]{article}

\usepackage{amsmath,amssymb}
\usepackage{iftex}
\ifPDFTeX
  \usepackage[T1]{fontenc}
  \usepackage[utf8]{inputenc}
  \usepackage{textcomp} 
\else 
  \usepackage{unicode-math}
  \defaultfontfeatures{Scale=MatchLowercase}
  \defaultfontfeatures[\rmfamily]{Ligatures=TeX,Scale=1}
\fi
\usepackage{lmodern}
\ifPDFTeX\else  
\fi
\IfFileExists{upquote.sty}{\usepackage{upquote}}{}
\IfFileExists{microtype.sty}{
  \usepackage[]{microtype}
  \UseMicrotypeSet[protrusion]{basicmath} 
}{}
\makeatletter
\@ifundefined{KOMAClassName}{
  \IfFileExists{parskip.sty}{%
    \usepackage{parskip}
  }{
    \setlength{\parindent}{0pt}
    \setlength{\parskip}{6pt plus 2pt minus 1pt}}
}{
  \KOMAoptions{parskip=half}}
\makeatother
\usepackage{xcolor}
\ifx\paragraph\undefined\else
  \let\oldparagraph\paragraph
  \renewcommand{\paragraph}[1]{\oldparagraph{#1}\mbox{}}
\fi
\ifx\subparagraph\undefined\else
  \let\oldsubparagraph\subparagraph
  \renewcommand{\subparagraph}[1]{\oldsubparagraph{#1}\mbox{}}
\fi

\providecommand{\tightlist}{%
  \setlength{\itemsep}{0pt}\setlength{\parskip}{0pt}}\usepackage{longtable,booktabs,array}
\usepackage{calc} 
\usepackage{etoolbox}
\makeatletter
\patchcmd\longtable{\par}{\if@noskipsec\mbox{}\fi\par}{}{}
\makeatother
\IfFileExists{footnotehyper.sty}{\usepackage{footnotehyper}}{\usepackage{footnote}}
\makesavenoteenv{longtable}
\usepackage{graphicx}
\makeatletter
\def\maxwidth{\ifdim\Gin@nat@width>\linewidth\linewidth\else\Gin@nat@width\fi}
\def\maxheight{\ifdim\Gin@nat@height>\textheight\textheight\else\Gin@nat@height\fi}
\makeatother
\setkeys{Gin}{width=\maxwidth,height=\maxheight,keepaspectratio}
\makeatletter
\def\fps@figure{htbp}
\makeatother
\newlength{\cslhangindent}
\newlength{\csllabelwidth}
\newlength{\cslentryspacingunit} 
\newenvironment{CSLReferences}[2] 
 {
  \setlength{\parindent}{0pt}
  \ifodd #1
  \let\oldpar\par
  \def\par{\hangindent=\cslhangindent\oldpar}
  \fi
  \setlength{\parskip}{#2\cslentryspacingunit}
 }%
 {}
\usepackage{calc}

\newcommand{\CSLLeftMargin}[1]{\parbox[t]{\csllabelwidth}{#1}}
\newcommand{\CSLRightInline}[1]{\parbox[t]{\linewidth - \csllabelwidth}{#1}\break}

\usepackage{arxiv}
\usepackage{orcidlink}
\usepackage{amsmath}
\usepackage[T1]{fontenc}
\makeatletter
\@ifpackageloaded{caption}{}{\usepackage{caption}}
\AtBeginDocument{%
\ifdefined\contentsname
  \renewcommand*\contentsname{Table of contents}
\else
  \newcommand\contentsname{Table of contents}
\fi
\ifdefined\listfigurename
  \renewcommand*\listfigurename{List of Figures}
\else
  \newcommand\listfigurename{List of Figures}
\fi
\ifdefined\listtablename
  \renewcommand*\listtablename{List of Tables}
\else
  \newcommand\listtablename{List of Tables}
\fi
\ifdefined\figurename
  \renewcommand*\figurename{Figure}
\else
  \newcommand\figurename{Figure}
\fi
\ifdefined\tablename
  \renewcommand*\tablename{Table}
\else
  \newcommand\tablename{Table}
\fi
}
\@ifpackageloaded{float}{}{\usepackage{float}}
\floatstyle{ruled}
\@ifundefined{c@chapter}{\newfloat{codelisting}{h}{lop}}{\newfloat{codelisting}{h}{lop}[chapter]}
\floatname{codelisting}{Listing}

\makeatother
\makeatletter
\@ifpackageloaded{caption}{}{\usepackage{caption}}
\@ifpackageloaded{subcaption}{}{\usepackage{subcaption}}
\makeatother
\makeatletter
\@ifpackageloaded{tcolorbox}{}{\usepackage[skins,breakable]{tcolorbox}}
\makeatother
\makeatletter
\@ifundefined{shadecolor}{\definecolor{shadecolor}{rgb}{.97, .97, .97}}
\makeatother
\ifLuaTeX
  \usepackage{selnolig}  
\fi
\IfFileExists{bookmark.sty}{\usepackage{bookmark}}{\usepackage{hyperref}}
\IfFileExists{xurl.sty}{\usepackage{xurl}}{} 
\hypersetup{
  pdftitle={Alternatives to the Scaled Dot Product for Attention in the Transformer Neural Network Architecture},
  pdfauthor={James Bernhard},
  colorlinks=true,
  linkcolor={blue},
  filecolor={Maroon},
  citecolor={Blue},
  urlcolor={Blue},
  pdfcreator={LaTeX via pandoc}}

\newcommand{\runninghead}{A Preprint }
\renewcommand{\runninghead}{Alternatives to the Scaled Dot Product }
\title{Alternatives to the Scaled Dot Product for Attention in the
Transformer Neural Network Architecture}
\author{\textbf{James
Bernhard}~\orcidlink{0009-0001-0762-2035}\\\\University of Puget
Sound\\\\\href{mailto:jbernhard@pugetsound.edu}{jbernhard@pugetsound.edu}}
\date{}
\begin{document}
\maketitle
\begin{abstract}
The transformer neural network architecture uses a form of attention in
which the dot product of query and key is divided by the square root of
the key dimension before applying softmax. This scaling of the dot
product is designed to avoid the absolute value of the dot products
becoming so large that applying softmax leads to vanishing gradients.

In this paper, we propose some alternative scalings, including dividing
the dot product instead by the sum of the key lengths before applying
softmax. We use simulated keys and queries to show that in many
situations this appears to be more effective at avoiding regions where
applying softmax leads to vanishing gradients.
\end{abstract}
\ifdefined\Shaded\renewenvironment{Shaded}{\begin{tcolorbox}[interior hidden, borderline west={3pt}{0pt}{shadecolor}, enhanced, frame hidden, boxrule=0pt, sharp corners, breakable]}{\end{tcolorbox}}\fi

Attention plays a prominent role in the transformer neural network
architecture, as indicated by the title of the landmark paper
introducing the architecture, ``Attention Is All You Need''
\protect\hyperlink{ref-Vaswani2017}{{[}1{]}}, by Vaswani et al.~The way
that attention is used in the transformer architecture builds on the way
attention was introduced by Bahdanau et al.
\protect\hyperlink{ref-Bahdanau2014}{{[}2{]}} and further developed by
Luong et al. \protect\hyperlink{ref-Luong2015}{{[}3{]}}.

In this paper, to explore how attention is used in the transformer
architecture we first describe attention in abstract terms. We then
discuss the problem that inspired the introduction of the scaled dot
product in ``Attention Is All You Need''
\protect\hyperlink{ref-Vaswani2017}{{[}1{]}}, which leads us to a shift
in perspective that suggests other possible ways of addressing that
problem. Next we propose dividing by the sum of the key lengths instead
of the square root of the key dimension, and we generate some simulated
queries and keys with independent standard normally distributed
components in order to compare the two methods. We also briefly discuss
some other possible scalings that might also help avoid vanishing
gradients.

\hypertarget{defining-attention}{%
\section{Defining attention}\label{defining-attention}}

A \emph{scalar attention function} is a function
\(a: \mathbb{R}^d \times \mathbb{R}^d \to \mathbb{R}\) for some \(d\).
The first argument of \(a\) is called a \(query\) and the second
argument a \(key\). For \(q, k \in \mathbb{R}^d\), we will call the real
number \(a(q, k)\) the \emph{scalar attention} of \(q\) on \(k\).

Given an attention function
\(a: \mathbb{R}^d \times \mathbb{R}^d \to \mathbb{R}\) and a finite
ordered set \(\mathcal{K}\) of keys
\(k_1, \ldots, k_n \in \mathbb{R}^d\), we can define a function
\(a_{\mathcal{K}}: \mathbb{R}^d \to \mathbb{R}^n\), which we will call
the \emph{vector attention function} associated with \(\mathcal{K}\) and
based on \(a\), by: \[
a_{\mathcal{K}}(q) =
\begin{bmatrix}
a(q, k_1)\\
a(q, k_2)\\
\vdots\\
a(q, k_n)
\end{bmatrix}.
\] We will call \(a_{\mathcal{K}}(q)\) the \emph{vector attention} of
\(q\) on \(\mathcal{K}\).

Suppose that, for this set \(\mathcal{K}\) of \(n\) keys, we have a
function \(s_{\mathcal{K}}: \mathbb{R}^n \to \Delta^{n-1}\), where
\(\Delta^{n-1}\) is the standard \((n-1)\)-simplex in \(\mathbb{R}^n\):
\[
\Delta^{n-1} = \{ \begin{bmatrix}
x_1\\
x_2\\
\vdots\\
x_n
\end{bmatrix}
\in \mathbb{R}^n\
|\
\sum_{i=1}^n x_i = 1\
\text{and}\
\forall i\
x_i \geq 0
\}.
\]

We will call such a function \(s_{\mathcal{K}}\) a \emph{rescaling
function}.

With a rescaling function, we can then define the \emph{rescaled vector
attention function} \(A_{\mathcal{K}}: \mathbb{R}^d \to \Delta^{n-1}\)
based on \(a\), \(\mathcal{K}\), and \(s_{\mathcal{K}}\) by:

\[
A_{\mathcal{K}} = s_{\mathcal{K}} \circ a_{\mathcal{K}}.
\]

In the context of ``Attention Is All You Need''
\protect\hyperlink{ref-Vaswani2017}{{[}1{]}}, the authors describe what
we have termed their scalar attention function as a ``scaled dot
product'': \[
a(q, k) = \frac{q \cdot k}{\sqrt{d}}.
\] For each set \(\mathcal{K}\) of keys \(k_1, \ldots, k_n\), they use
\(\text{softmax}_n: \mathbb{R}^n \to \Delta^{n-1}\) as what we have
called a rescaling function, defined by: \[
\text{softmax}_n(
\begin{bmatrix}
x_1\\
x_2\\
\vdots\\
x_n
\end{bmatrix}
) =
\frac{1}{\sum_{i=1}^n \exp(x_i)}
\begin{bmatrix}
\exp(x_1)\\
\exp(x_2)\\
\vdots\\
\exp(x_n)
\end{bmatrix}.
\]

This produces the following rescaled vector attention function
\(A_{\mathcal{K}}: \mathbb{R}^d \to \mathbb{R}^n\): \[
A_{\mathcal{K}}(q) = 
\frac{1}{\sum_{i=1}^n \exp(q \cdot k_i/\sqrt{d})}
\begin{bmatrix}
\exp(q \cdot k_1/\sqrt{d})\\
\exp(q \cdot k_2/\sqrt{d})\\
\vdots\\
\exp(q \cdot k_n/\sqrt{d})
\end{bmatrix}.
\]

Their paper also includes some additional aspects of attention that are
important for its implementation in the transformer architecture, but
that we mention only briefly here:

\begin{enumerate}
\def\labelenumi{\arabic{enumi}.}
\tightlist
\item
  In addition to keys \(k_1, \ldots, k_n \in \mathbb{R}^d\), they also
  have a finite ordered set of \emph{values}
  \(v_1, \ldots, v_n \in \mathbb{R}^t\). (Since keys and values are to
  be thought of as pairs, the number of keys and values must be the
  same. However, the dimensions of the spaces where they reside may be
  different.) To arrive at what they term their ``Attention'' function,
  they first apply the above rescaled vector attention function and then
  apply a linear transformation \(T: \mathbb{R}^n \to \mathbb{R}^t\)
  (restricted to the domain \(\Delta_{n-1}\)) given by \[
  T(
  \begin{bmatrix}
  x_1\\
  x_2\\
  \vdots\\
  x_n
  \end{bmatrix}
  ) =
  \sum_{i=1}^n x_i v_i.
  \] This produces a vector in the convex hull of \(v_1, \ldots, v_n\).
\item
  Also, they write their ``Attention'' function in terms of a \emph{set}
  of (row vector) queries \(q_1, \ldots, q_m \in \mathbb{R}^d\) that
  form the rows of a matrix, rather than in terms of individual queries.
\end{enumerate}

These aspects are important for the implementation of attention in the
transformer architecture but not for our discussion here.

\hypertarget{compensating-for-the-dimension-of-the-keys}{%
\section{Compensating for the dimension of the
keys}\label{compensating-for-the-dimension-of-the-keys}}

The authors of ``Attention Is All You Need''
\protect\hyperlink{ref-Vaswani2017}{{[}1{]}} explain their reason for
dividing the dot product by \(\sqrt{d}\) (or \(\sqrt{d_k}\) in their
notation) in what we have termed their scalar attention function: ``We
suspect that for large values of \(d_k\), the dot products grow large in
magnitude, pushing the softmax function into regions where it has
extremely small gradients. To counteract this effect, we scale the dot
products by \(1/\sqrt{d_k}\).'' They elaborate further in a footnote
that the issue can be observed by noting that if the components of the
queries and keys are independent random variables with mean \(0\) and
variance \(1\), then their dot products have mean \(0\) and variance
\(d\).

Their explanation can be further illustrated through some simulations,
and understanding these simulations will also be helpful when we use
similar simulations to compare alternative rescalings later in this
paper.

\begin{figure}

{\centering \includegraphics{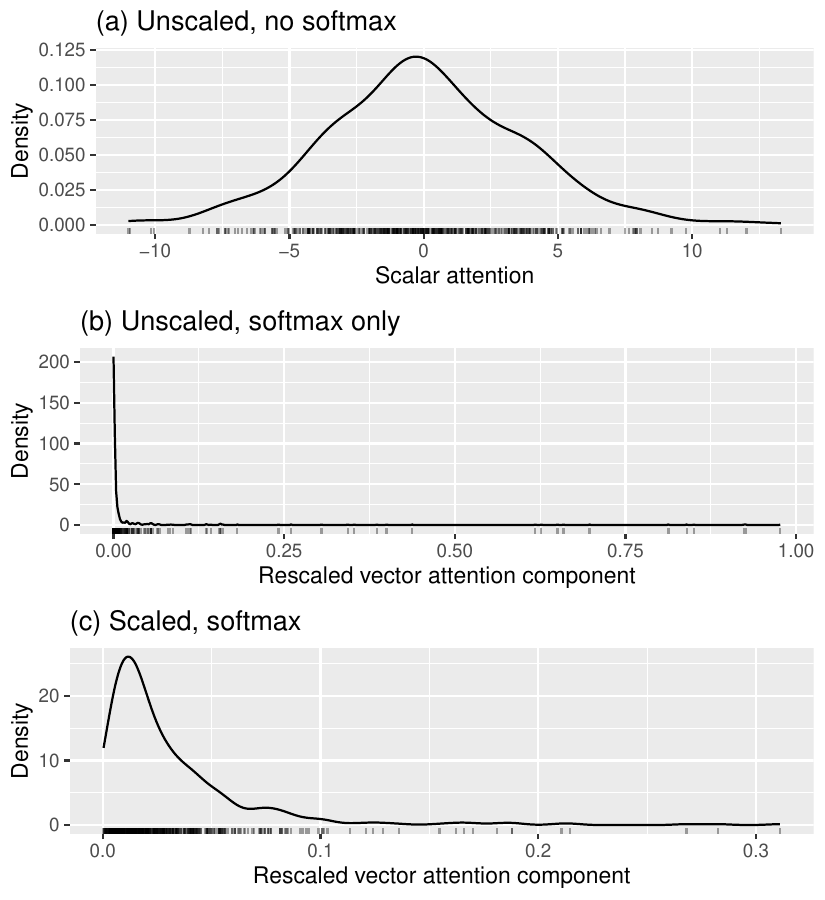}

}

\caption{\label{fig-sqrt-vs-unscaled}The first component of vector
attention for 500 simulated queries. (a) These points have not had any
scaling, and softmax has not been applied. They are simply dot products
of queries with the first key. (b) These dot products of queries with
the first key have not been scaled, but they have had softmax applied.
(c) These dot products of queries with the first key have been divided
by the square root of the key dimension and have then had softmax
applied.}

\end{figure}

For each of the plots in Figure~\ref{fig-sqrt-vs-unscaled}, we simulate
all the components of every query and key as being independent and
standard normally distributed. We generate a set of \(32\) keys, and we
give them a very small dimension of \(16\). (The effect we are
illustrating is so large that even such a small dimension shows it
strongly.) To compute attentions, we generate \(500\) queries, also each
of dimension \(16\). We then make a kernel density estimate plot of just
the first component of the rescaled vector attention function applied to
all the queries. By symmetry, the other components should look similar,
and this approach avoids any issues with displaying (in the same kernel
density estimate plot) multiple components of the rescaled vector
attention, which lack independence.

In these three plots, each point represents the following:

\begin{enumerate}
\def\labelenumi{\alph{enumi}.}
\tightlist
\item
  In (a), a single unscaled dot product.
\item
  In (b), a single unscaled dot product after softmax has been applied
  to it.
\item
  In (c), a single scaled dot product (scaled by \(1/\sqrt{d}\)) after
  softmax has been applied to it.
\end{enumerate}

Comparing the three plots in Figure~\ref{fig-sqrt-vs-unscaled}, we can
see how softmax, with and without scaling, distorts the shape of the
original distribution of unscaled dot products.

Even with such a small key dimension as \(16\),
Figure~\ref{fig-sqrt-vs-unscaled}(b) shows that applying softmax without
scaling severely distorts the shape of the distribution of the original
dot products. Because the shape after applying softmax is weighted
heavily near \(0\) and has a pronounced right skew, it is apparent that
applying softmax to the unscaled dot product would be prone to vanishing
gradient issues.

Although Figure~\ref{fig-sqrt-vs-unscaled}(c) shows a mild right skew
that was not present in the distribution before softmax was applied,
from the plot it is clear that dividing the dot products by \(\sqrt{d}\)
before applying softmax did indeed help preserve the general shape of
the distribution compared to not scaling at all. Correspondingly, this
method helps alleviate the vanishing gradient issue with softmax (as has
also been demonstrated time and again in its widespread use).

However, there are two things to think about here:

\begin{enumerate}
\def\labelenumi{\arabic{enumi}.}
\item
  The choice to divide by \(\sqrt{d}\) was based on the absolute values
  of the query and key dot products growing too large because of their
  dimension, but there are other reasons besides the dimension why the
  absolute value of these dot products might grow large. It would be
  good to compensate for those too.
\item
  The reason why we want to compensate for the size of the dot products
  is that we are planning to apply softmax, but notice that it is
  impossible to apply softmax without knowing how many keys are in
  \(\mathcal{K}\). As such, we might not be able to avoid all of the
  softmax-related difficulties by adjusting the scalar attention
  function, since this function has nothing to do with \(\mathcal{K}\),
  only with individual keys. Instead we might want to adjust the
  rescaling function, which does depend on the choice of
  \(\mathcal{K}\).
\end{enumerate}

These considerations lead us to rethink the way we view dividing by
\(\sqrt{d}\) in the transformer neural network architecture. Instead of
thinking of the ``scaled dot product'' \((q \cdot k)/\sqrt{d}\) as the
scalar attention function and softmax as the rescaling function, we view
the scalar attention function as the ordinary dot product \(q \cdot k\)
and the rescaling function as what might be called a ``prescaled softmax
function'' \(s_{\mathcal{K}}: \mathbb{R}^n \to \Delta_{n-1}\): \[
s_{\mathcal{K}}(
\begin{bmatrix}
x_1\\
x_2\\
\vdots\\
x_n
\end{bmatrix}
) =
\frac{1}{\sum_{i=1}^n \exp(x_i/\sqrt{d})}
\begin{bmatrix}
\exp(x_1/\sqrt{d})\\
\exp(x_2/\sqrt{d})\\
\vdots\\
\exp(x_n/\sqrt{d})
\end{bmatrix}.
\]

This minor shift in perspective is unimportant for using the
``Attention'' function directly as formulated in ``Attention Is All You
Need'' \protect\hyperlink{ref-Vaswani2017}{{[}1{]}}. However, for our
purposes here it is crucial. It suggests that the ordinary, unscaled dot
product (as used already in \protect\hyperlink{ref-Luong2015}{{[}3{]}})
is actually a good scalar attention function; rescaling follows it only
because of how it is to be used in a particular neural network context.
This might seem like a small change, but it opens up the possibility
that the rescaling might depend not solely on the dimension of the keys
(as with dividing by \(\sqrt{d}\)) but possibly on the number of keys in
\(\mathcal{K}\) (as softmax does), or even on the keys
\(k_1, \ldots, k_n\) themselves.

\hypertarget{exploring-rescalings}{%
\section{Exploring rescalings}\label{exploring-rescalings}}

In light of the above considerations, we take the following as our
starting point: the ordinary, unscaled dot product is a good scalar
attention function for the transformer architecture, and our task is to
figure out a suitable rescaling function
\(s_{\mathcal{K}}: \mathbb{R}^n \to \Delta_{n-1}\) that preserves as
much of the overall shape of the distribution of the unscaled dot
products as possible.

Note that the rescaling function doesn't interact with queries directly,
so in this approach, it can't be helped if \(\| q \|\) gets large and so
makes the dot product large. If we were modifying the scalar attention
function, it might be tempting to divide it by \(\| q \|\), but (even
ignoring the possibility of this being \(0\)) that would make the scalar
attention function independent of \(\| q \|\), which seems undesirable.
It might be worth thinking about whether anything different should be
done with \(q\) in the scalar attention function, but for our purposes
here we are leaving \(q\) out of the investigation and focusing solely
on finding a rescaling function.

Suppose then that we have the dot product as our scalar attention
function \(a: \mathbb{R}^d \times \mathbb{R}^d \to \mathbb{R}\): \[
a(q, k) = q \cdot k.
\] Also suppose that we are given a finite ordered set \(\mathcal{K}\)
of keys \(k_1, \ldots, k_n \in \mathbb{R}^d\). We can then define the
vector attention function
\(a_{\mathcal{K}}: \mathbb{R}^d \times \mathbb{R}^d \to \mathbb{R}^n\)
by: \[
a_{\mathcal{K}}(q) =
\begin{bmatrix}
q \cdot k_1\\
q \cdot k_2\\
\vdots\\
q \cdot k_n
\end{bmatrix}.
\] Our task is to find a suitable rescaling function
\(s_{\mathcal{K}}: \mathbb{R}^n \to \Delta_{n-1}\).

A natural idea for a rescaling function would be
\(\text{softmax}_n: \mathbb{R}^n \to \Delta_{n-1}\), but as discussed
above and in ``Attention Is All You Need''
\protect\hyperlink{ref-Vaswani2017}{{[}1{]}}, this function by itself is
prone to vanishing gradients because the absolute values of the dot
products that make up each component can easily get very large as the
dimension \(d\) of the keys and queries does. The authors of ``Attention
Is All You Need'' \protect\hyperlink{ref-Vaswani2017}{{[}1{]}} suggest
pre-dividing by \(\sqrt{d}\) before applying softmax, which works well
for avoiding this problem.

However, as also discussed above, there are other reasons why the
absolute values of the dot products might get large besides just the
dimension of the key space. Keeping in mind that the function
\(s_{\mathcal{K}}\) can't depend on the queries but that it can depend
on \(k_1, \ldots, k_n\), some thought and experimentation suggests
pre-dividing by \(k_{total}\), the sum of the key lengths: \[
k_{total} = \sum_{i=1}^n \|k_i\|.
\] (Note that \(k_{total}\) should indeed be nonzero in all practical
applications.) In other words, as a rescaling function
\(s_{\mathcal{K}}: \mathbb{R}^n \to \Delta_{n-1}\), we propose: \[
s_{\mathcal{K}}(
\begin{bmatrix}
x_1\\
x_2\\
\vdots\\
x_n
\end{bmatrix}
) =
\frac{1}{\sum_{i=1}^n \exp(x_i/k_{total})}
\begin{bmatrix}
\exp(x_1/k_{total})\\
\exp(x_2/k_{total})\\
\vdots\\
\exp(x_n/k_{total})
\end{bmatrix}.
\]

Some of the intuition behind this is that, by the Cauchy-Schwarz
inequality, \[
\frac{ |k_i \cdot q| }{k_{total}} \leq \frac{ \|k_i\| }{k_{total}} \|q\| \leq \| q \|.
\] As discussed earlier, the rescaling function doesn't interact
directly with the queries themselves. But the previous inequality shows
that as long as the lengths of the queries don't get too large, then the
relative sizes of the components of the vectors to which softmax is
applied won't be distorted too much.

\hypertarget{some-simulations-to-compare-rescalings}{%
\section{Some simulations to compare
rescalings}\label{some-simulations-to-compare-rescalings}}

We now use simulations to compare this rescaling function to the one
used (with a slightly different point of view, as discussed above) in
``Attention Is All You Need''
\protect\hyperlink{ref-Vaswani2017}{{[}1{]}}. Both are of the following
form:

\[
s_{\mathcal{K}}(
\begin{bmatrix}
x_1\\
x_2\\
\vdots\\
x_n
\end{bmatrix}
) =
\frac{1}{\sum_{i=1}^n \exp(x_i/c_{\mathcal{K}})}
\begin{bmatrix}
\exp(x_1/c_{\mathcal{K}})\\
\exp(x_2/c_{\mathcal{K}})\\
\vdots\\
\exp(x_n/c_{\mathcal{K}})
\end{bmatrix},
\] and \(c_{\mathcal{K}}\) equals either \(\sqrt{d}\) or \(k_{total}\).

\begin{figure}

{\centering \includegraphics{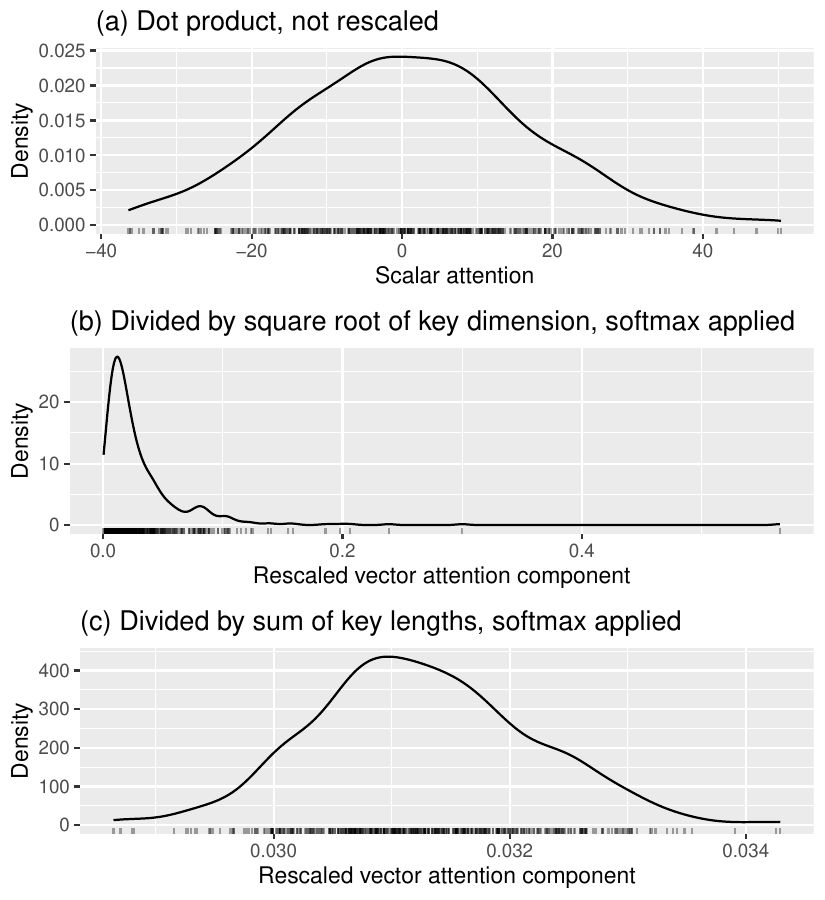}

}

\caption{\label{fig-rescaling-comparison}The first component of vector
attention for 500 simulated queries. (a) These scalar attentions have
not had any rescaling. They are simply dot products of queries with the
first key. (b) These are the first component of the rescaled vector
attention, where the rescaling is dividing by the square root of the key
dimension and then applying softmax. (c) These are the first component
of the rescaled vector attention, where the rescaling is dividing by the
sum of the key lengths and then applying softmax.}

\end{figure}

In each of the plots in Figure~\ref{fig-rescaling-comparison}, we
simulate all the components of every query and key as being independent
and standard normally distributed. We generate a set of \(32\) keys,
each of dimension \(256\). To compute the vector attentions, we generate
\(500\) queries, also each of dimension \(256\). We then make a kernel
density estimate plot of just the first component of either the vector
attention function or the rescaled vector attention function applied to
all the queries. By symmetry, the other components should look similar,
and as before, this approach avoids any issues with displaying (in the
same kernel density estimate plot) multiple components of the rescaled
vector attention, which lack independence.

Each point in the three plots in Figure~\ref{fig-rescaling-comparison}
represents:

\begin{enumerate}
\def\labelenumi{\alph{enumi}.}
\tightlist
\item
  In (a), a single unscaled dot product.
\item
  In (b), a dot product which has been rescaled by the above
  \(s_{\mathcal{K}}\) with \(c_{\mathcal{K}} = \sqrt{d}\), as in
  ``Attention Is All You Need''
  \protect\hyperlink{ref-Vaswani2017}{{[}1{]}},
\item
  In (c), a dot product which has been rescaled by the above
  \(s_{\mathcal{K}}\) with
  \(c_{\mathcal{K}} = k_{total} = \sum_{i=1}^n \|k_i\|\).
\end{enumerate}

The axes on all three plots are on completely different scales, but we
have displayed them in this way (using the default scales chosen by the
computer) in order to be able to compare the shapes of the
distributions. In these plots, it is apparent that pre-dividing by
\(k_{total}\) before applying softmax preserves the shape of the
distribution much better than pre-dividing by \(\sqrt{d}\).

These simulations have used an independent standard normal distribution
for each component of the queries and keys. Further experimentation
readily shows that when normal distribution means and variances change
(and even more so when distribution families change), pre-dividing by
\(\sqrt{d}\) leads to greater distortions of the distribution shape than
pre-dividing by \(k_{total}\). Also, these patterns persist even when
\(d\) and \(n\) vary over a wide range.

To explore these variations further, the R package that the author wrote
to generate the simulations and plots for this paper can be used. It is
available at \url{https://github.com/James-Bernhard/attnsims}.

\hypertarget{further-considerations}{%
\section{Further considerations}\label{further-considerations}}

Once we view dividing the dot product by \(\sqrt{d}\) in the transformer
architecture as part of the rescaling function rather than as a
modification of the dot product in the scalar attention function, many
other rescaling possibilities arise. In our explorations, pre-dividing
by \(k_{total}\) has appeared to decrease the distortion by the softmax
function and help avoid the issue of vanishing gradients, but some other
possibilities that might also be considered are:

\begin{itemize}
\tightlist
\item
  dividing by \(\sum_{i=1}^n \| k_i \|/n\),
\item
  dividing by \(\sqrt{\sum_{i=1}^n \| k_i \|^2}\),
\item
  more generally, dividing by
  \(\left(\sum_{i=1}^n \| k_i \|^p \right)^{1/p}\).
\end{itemize}

These are more complicated, and in the author's explorations, none of
them seemed to perform as well as dividing by \(k_{total}\). But perhaps
there are other possibilities that do.

It is also possible that different rescalings might be more effective in
different contexts. As such, it would be good to consider rescaling
functions other than just the ones described here. If it is feasible,
multiple possible rescaling functions might even be tested in the
specific context in which they are to be applied.

It is worth noting that dividing by \(n \sqrt{d}\) seems to exhibit
similar behavior to dividing by \(k_{total}\) for independent standard
normally distributed key and query components. If it is impractical to
compute \(k_{total}\) in a particular situation, perhaps dividing by
\(n \sqrt{d}\) could be substituted for it. As might be expected though,
dividing by \(n \sqrt{d}\) doesn't appear to behave as well for
distributions other than standard normal.

\hypertarget{conclusion}{%
\section{Conclusion}\label{conclusion}}

Vanishing gradients arise when individual components resulting from
applying softmax approach \(0\) or \(1\), and this issue can be seen in
the distortion of the shapes of the distributions of the vector
attention components. Dividing by the square root of the key dimension
alleviates this distortion well, but, judging from simulations, in many
situations dividing by the sum of the key lengths appears to preserve
the shape of the distribution even better.

Of course, the real test of the utility of dividing by the sum of the
key lengths lies in how it performs in its intended setting in the
transformer neural network architecture. Although further and more
systematic testing is still needed, the author has conducted preliminary
experiments using this modified transformer architecture and it appears
to perform well.

The simulations and plots in this paper were generated using R version
4.3.0 \protect\hyperlink{ref-R2023}{{[}4{]}} and the R packages
\texttt{ggplot2} \protect\hyperlink{ref-ggplot22016}{{[}5{]}} and
\texttt{rlang} \protect\hyperlink{ref-rlang2023}{{[}6{]}}. The R package
that the author wrote to generate the simulations and plots in this
paper is available at \url{https://github.com/James-Bernhard/attnsims}.

\hypertarget{bibliography}{%
\section*{References}\label{bibliography}}
\addcontentsline{toc}{section}{References}

\hypertarget{refs}{}
\begin{CSLReferences}{0}{0}
\leavevmode\vadjust pre{\hypertarget{ref-Vaswani2017}{}}%
\CSLLeftMargin{{[}1{]} }%
\CSLRightInline{A. Vaswani \emph{et al.}, {``Attention is all you
need.''} 2017. doi:
\href{https://doi.org/10.48550/arXiv.1706.03762}{10.48550/arXiv.1706.03762}.
Available: \url{https://arxiv.org/abs/1706.03762}}

\leavevmode\vadjust pre{\hypertarget{ref-Bahdanau2014}{}}%
\CSLLeftMargin{{[}2{]} }%
\CSLRightInline{D. Bahdanau, K. Cho, and Y. Bengio, {``Neural machine
translation by jointly learning to align and translate.''} 2014. doi:
\href{https://doi.org/10.48550/arXiv.1409.0473}{10.48550/arXiv.1409.0473}.
Available: \url{https://arxiv.org/abs/1409.0473}}

\leavevmode\vadjust pre{\hypertarget{ref-Luong2015}{}}%
\CSLLeftMargin{{[}3{]} }%
\CSLRightInline{M.-T. Luong, H. Pham, and C. D. Manning, {``Effective
approaches to attention-based neural machine translation.''} 2015. doi:
\href{https://doi.org/10.48550/arXiv.1508.04025}{10.48550/arXiv.1508.04025}.
Available: \url{https://arxiv.org/abs/1508.04025}}

\leavevmode\vadjust pre{\hypertarget{ref-R2023}{}}%
\CSLLeftMargin{{[}4{]} }%
\CSLRightInline{R Core Team, \emph{R: A language and environment for
statistical computing}. Vienna, Austria: R Foundation for Statistical
Computing, 2023. Available: \url{https://www.R-project.org/}}

\leavevmode\vadjust pre{\hypertarget{ref-ggplot22016}{}}%
\CSLLeftMargin{{[}5{]} }%
\CSLRightInline{H. Wickham, \emph{ggplot2: Elegant graphics for data
analysis}. Springer-Verlag New York, 2016. Available:
\url{https://ggplot2.tidyverse.org}}

\leavevmode\vadjust pre{\hypertarget{ref-rlang2023}{}}%
\CSLLeftMargin{{[}6{]} }%
\CSLRightInline{L. Henry and H. Wickham, \emph{Rlang: Functions for base
types and core r and 'tidyverse' features}. 2023. Available:
\url{https://CRAN.R-project.org/package=rlang}}

\end{CSLReferences}

\end{document}